# Lessons Learned from Evaluation of LLM based Multi-agents in Safer Therapy Recommendation

Yicong Wu[1]
21918142@zju.edu.cn

Ting Chen[2]
ct010151452@zju.edu.cn

Irit Hochberg[3,4]
iritho@hymc.gov.il

Zhoujian Sun[5]
sunzhoujian.szj@antgroup.com

Ruth Edry[3,6]
ruedry@gmail.com

Zhengxing Huang[1*]
zhengxinghuang@zju.edu.cn

Mor Peleg[7*]
morpeleg@is.haifa.ac.il

[1] Zhejiang University, Hangzhou, China
[2] The First Affiliated Hospital, Zhejiang University School of Medicine, Hangzhou, China
[3] Bruce Rappaport Faculty of Medicine, Technion – Israel Institute of Technology
[4] Hillel Yaffe Medical Center, Hadera, Israel
[5] Ant Group, Hangzhou, China
[6] Rambam Medical Center, Haifa, Israel
[7] Department of Information Systems, University of Haifa, Haifa, Israel

*Abstract*—Therapy recommendation for chronic patients with multimorbidity is challenging due to risks of treatment conflicts. Existing decision support systems face scalability limitations. Inspired by the way in which general practitioners (GP) manage multimorbidity patients, occasionally convening multidisciplinary team (MDT) collaboration, this study investigated the feasibility and value of using a Large Language Model (LLM)-based multi-agent system (MAS) for safer therapy recommendations. We designed a single agent and a MAS framework simulating MDT decision-making by enabling discussion among LLM agents to resolve medical conflicts. The systems were evaluated on therapy planning tasks for multimorbidity patients using benchmark cases. We compared MAS's performance with single-agent approaches and real-world benchmarks. An important contribution of our study is the definition of evaluation metrics that go beyond the technical precision and recall and allow the inspection of clinical goals met and medication burden of the proposed advices to a gold standard benchmark. Our results show that with current LLMs, a single agent GP performs as well as MDTs. The best-scoring models provide correct recommendations that address all clinical goals, yet the advices are incomplete. Some models also present unnecessary medications, resulting in unnecessary conflicts between medication and conditions or drug-drug interactions. The benchmark dataset is available at https://github.com/MGCDS/benchmarkframework. The code, prompts, case dialogue results are available at https://github.com/aiformed/MAS_for_Therapy_Recom.

## I. Introduction

Therapy recommendation becomes particularly challenging in chronic multimorbidity scenarios, where patients suffer from multiple coexisting conditions [21]. In such cases, clinicians must prescribe several medications concurrently, increasing the risk of drug-drug interactions (DDIs), redundant prescriptions, and drug-condition contraindications [22]. Detecting and mitigating these **conflicts** is crucial to ensure patient safety, and while no solution can be entirely conflict-free, clinical decision-support systems (CDSS) should prioritize the most severe conflicts.

Some existing multimorbidity therapy recommendation systems rely on manual encoding of clinical guidelines into computer-interpretable formats, facing significant scalability challenges [1]. Other systems adopt deep neural network-based approaches, but often suffer from a lack of transparency and explainability [7-10]. A potential approach that offers both scalability and interpretability is the use of Large Language Models (LLMs). LLMs have demonstrated significant potential in various medical applications [11-13]. Among these, diagnosis and therapy recommendations are two critical aspects of clinical decision support. However, while diagnostic capabilities of LLMs have received increasing attention [16, 18, 20], research on LLM-based therapy recommendation remains relatively limited.

In clinical practice, a multidisciplinary team (MDT) is often established when facing complex patient conditions, so as to integrate diverse clinical expertise and develop comprehensive treatment plans. Inspired by this MDT decision-making model, recent research has explored multi-agent systems (MAS) composed of specialized LLM agents to simulate collaborative reasoning [23]. These systems have shown promise in addressing complex diagnostic challenges [20]. However, the application of LLM-based MAS for therapy recommendation remains underexplored.

To test the hypothesis that a collaborative team of domain-specialized LLM agents can enhance conflict mitigation in therapy recommendation, we propose an MAS framework designed to detect and resolve conflicts, and compare its performance against both single-agent systems and real-world baselines. This investigation aims to assess the feasibility, advantages, and limitations of LLM-based MAS for therapy recommendation, and to derive actionable insights for their future development and integration into practice. The main contributions are as follows:

1. We propose a dynamically generated multi-agent framework that simulates real-world multidisciplinary expert consultations. The MAS replicates the multi-step workflow of MDTs as performed by clinicians in real-world settings, allowing LLMs to propose improved treatment plans by detecting and resolving conflicts.

2. We apply the proposed MAS to clinical benchmark cases to assess its completeness and correctness using detailed qualitative research. Results demonstrated that, compared to single-agent systems, the MAS showed promise in reducing conflicts while maintaining an appropriate medication count, and it is more closely aligning with real-world expert judgment for multimorbidity patients.

3. This study develop a new interpretable evaluation strategy. Unlike traditional aggregate metrics, we introduce specific improvement ratio that compares LLM-proposed treatment plans with original plans, focusing on conflict reduction and medication burden, similar to a hazard ratio.

## II. RELATED WORKS

### A. Knowledge-based Multimorbidity Recommendation System

Early multimorbidity clinical decision support approaches heavily rely on knowledge-intensive systems that translate clinical guidelines into formal representations [1]. These computer-interpretable guidelines (CIGs)-based systems leverage explicit knowledge representation (e.g., ontologies, logical rules) to solve complex clinical tasks. However, such approaches suffer from critical limitations in scalability and adaptability. For instance, GLARE-SSCPM requires ontology-based annotation of CIG tasks with drug categories and treatment intentions [2], while CIG Integration Ontology (CIG-IntO) approach depends on hand-crafted integration policies to resolve conflicts across multiple guidelines [3]. Similarly, systems like GoCom require manual curation of SNOMED-CT or NDF-RT concepts to annotate clinical goals within PROforma CIGs [4].

Overall, these systems offer strong explainability and high safety, while requiring extensive manual effort to encode clinical knowledge into structured formats, heavily relying on the domain experts. This lack of scalability limits their applicability in dynamic or large-scale clinical environments. This gap has motivated the shift toward more flexible, data-driven AI approaches, which aim to reduce manual overhead.

### B. Data-driven Medication Recommendation System

Traditional machine learning methods were used to reduce medication errors by enabling automated prescription verification, real-time detection of drug interactions and dosage issues, and supporting clinicians with evidence-based recommendations tailored to patient profiles [5,6].

More advanced systems powered by deep learning, such as reinforcement learning [7], molecular graph learning [8] and knowledge graph-based models [9], show great potential in improving medication safety. But the inherent "black-box" nature of these models often conflicts with the need for transparency and explainability in clinical practice [10], underscoring the urgent demand for more interpretable AI solutions. To bridge this gap, LLMs offer an unprecedented opportunity by providing human-readable rationales that enhance transparency and trust in AI recommendations [11-13].

Recent studies have shown that models like GPT-4 can align reasonably well with guidelines in initial diagnostics and treatment planning. Nevertheless, they exhibit significant gaps in areas such as surgical planning and precise application of guidelines [14]. Other evaluations have revealed that while GPT exhibits high diagnostic and medication recommendations accuracy across simulated cases, it also produces a concerning level (85%) of inappropriate medication suggestions, including omitting recommended medications or recommending unnecessary or potentially harmful prescriptions [15].

### C. Medical Multi-agent System

To overcome the limitations of single-agent LLMs in complex clinical scenarios, recent research has explored LLM-based integration methods, particularly the multi-agent systems. That is, multiple LLM-powered agents are assigned specialized roles and collaborate to collectively formulate and validate recommendations. This architecture mirrors the dynamics of real-world MDTs. Prior research has demonstrated that MAS can outperform single-agent systems in complex tasks. Studies like AI Hospital [16], ColaCare [17], MedAide [18], and MDAgents [19], diagnostic MAS [20], have shown the benefits of MAS in healthcare, including the ability to simulate diverse expert viewpoints and foster

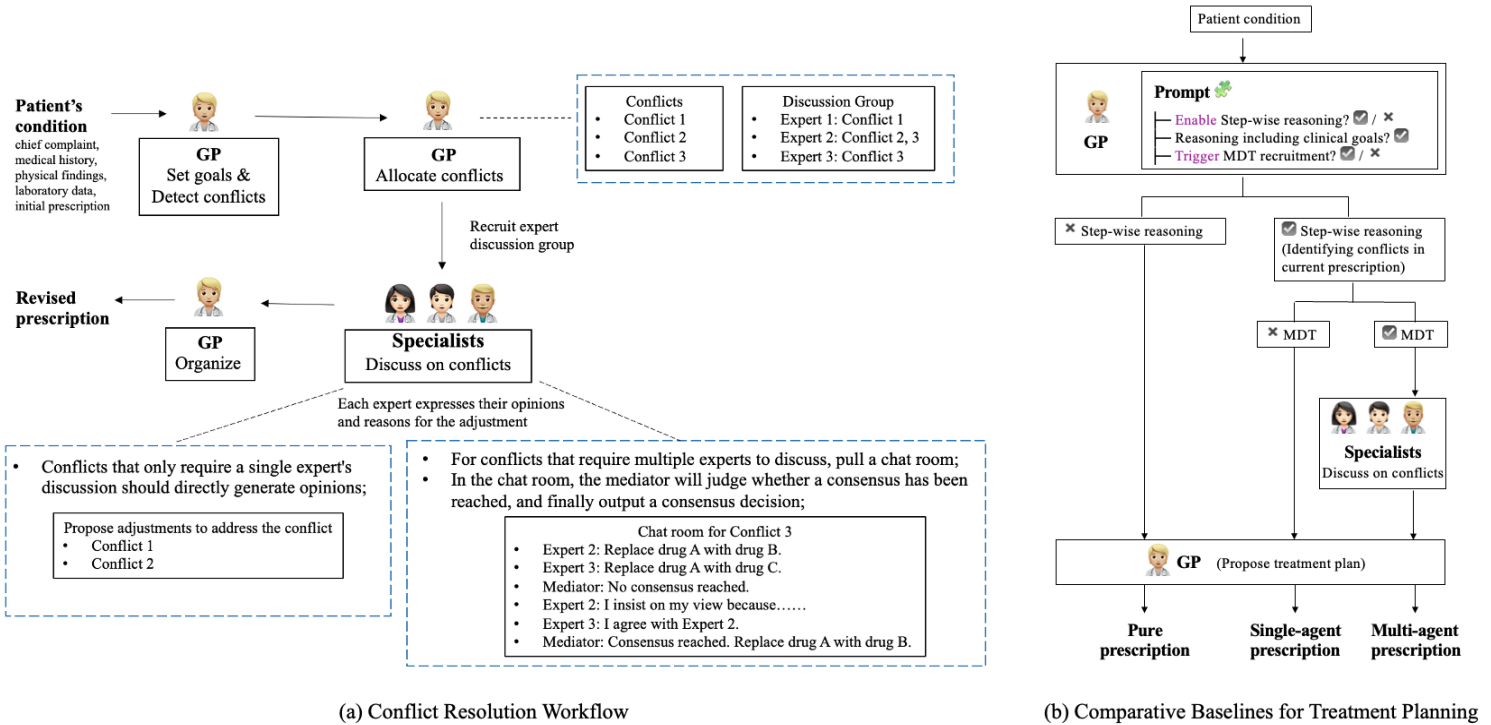

Fig. 1. Multi-agent Framework Design. (a) Conflict resolution workflow with multi-agent collaboration for medication recommendation. (b) Overview of comparative baselines for treatment planning, three experimental baselines are used to evaluate the proposed MAS for treatment planning.

collaborative or competitive agent interactions that improve solution reliability.

However, these prior works have largely focused on diagnosis and Q&A tasks, with limited attention to disease management, especially for multimorbidity patients, where conflicts are common due to multiple conditions and multiple drugs. In such contexts, treatment planning demands both accurate knowledge and conflict resolution mechanisms that balance conflicts and clinical goals.

To bridge this gap, we explore a novel application of LLM-based MAS collaboration in the domain of therapy recommendation. Unlike prior systems, our framework focuses on the resolution of conflicts, providing an interpretable, traceable workflow and validating final decisions using safety metrics. This design aims to enhance both the clinical transparency and the real-world utility of AI-assisted medication planning, particularly in scenarios requiring multidisciplinary reasoning.

## III. METHODS

### A. Core Conflict Resolution Workflow

Fig. 1a illustrates the conflict resolution workflow for a chronic patient with multimorbidity. Through prompt engineering, the LLM is assigned distinct roles to simulate interactions among multiple agents. The system operates with a primary role, the General Practitioner (GP), who oversees and coordinates the entire workflow.

The process begins with the GP reviewing the patient's clinical condition, including the chief complaint, medical history, physical findings, laboratory data, and initial prescription. Then the GP is asked to set clinical goals to manage the conditions and complaint, identify existing medications for the goals, and identify potential treatment **conflicts**, including DDIs and condition-specific contraindications that may increase patient risk:

$$C = G(P_0) \quad (1)$$

Where $C$ represents the detected conflict set, $P_0$ is the patient's clinical condition.

Any identified conflicts are reported back to the GP, who evaluates the information and convenes a MDT when he thinks that some aspects of the patient's condition are outside his expertise, selecting specialists relevant to each conflict for resolution.

During the conflict resolution phase, each specialist is assigned a subset of discussion points based on the identified conflicts. Some conflicts require input from multiple specialists, while others may be addressed by a single expert. For multi-expert conflicts, a collaborative forum is set up where each specialist presents their perspective, followed by rounds of peer evaluation. If disagreements arise, new arguments are proposed until a consensus is reached (please refer to Github for more details). The discussion proceeds in rounds $j \in [1, m]$. The maximal number of rounds m was set empirically to 5, based on observations that most meaningful disagreements or suggestions occur within the first few exchanges. Beyond this round, agents tend to repeat previously stated positions. Additionally, longer discussions have risk in exceeding the model's token limit. If a consensus is not reached within m rounds, a mediator agent is invoked to summarize the arguments and propose a consensus recommendation. In cases where only one expert is needed, the assigned specialist directly provides a recommendation. Upon resolution of all conflicts, the GP synthesizes the recommendations and finalizes the treatment plan, resulting in a revised prescription.

$$R_k = \begin{cases} Consensus(\{E_j(MDT_k, C_k)\}_{j=1}^m), |MDT_k| > 1 \\ E_1(S_i, C_k) \qquad\qquad\qquad\qquad , |MDT_k| = 1 \end{cases} \quad (2)$$

$$P_{revised} = Integrate(P_0, \sum_{k=1}^{K} R_k) \quad (3)$$

Where $R_k$ represents resolution outcome for conflict $C_k$, $MDT_k$ represents the subset of specialists assigned to $C_k$, $S_i$ is the $i$-th specialist, $E_j$ is the evaluation function in the $j$-th round of discussion, $m$ is the number of consensus rounds, $K$ is the total number of conflicts.

### B. Comparison baseline

To evaluate the generalizability of the system across different LLMs, we conducted experiments using four distinct models. In addition, we designed two control groups to assess the effectiveness of our MAS, as shown in Fig. 1b.

In the pure setting (first control), a single GP directly proposed a treatment plan based on the patient's condition, without any intermediate reasoning steps. The output is referred to as the Pure prescription.

In the single-agent setting (second control), the GP followed a step-by-step reasoning process, it first identified clinical goals and relevant medications from the current plan to these goals, and then detected potential conflicts before generating a treatment plan. The resulting output is referred to as the Single-agent prescription.

In the multi-agent setting, the GP performed the same initial reasoning steps as in the single-agent setting. However, after conflicts were detected, the GP selectively formed a MDT based on the complexity and nature of the conflicts. Those conflicts requiring specialists were assigned to relevant experts, in accordance with the constraint of minimizing MDT size. The MDT then resolved these issues through collaborative conflict-resolution process described earlier. Those conflict-specific recommendations were then sent to the GP, who reviewed and incorporated them into the final Multi-agent prescription, prioritizing goal consistency and conflict resolution.

### C. Model Selection and Settings

All experiments were conducted in a Python 3.12.9 environment. To simulate LLM-based role-playing using prompts, we utilized four distinct LLMs: GPT-4o [24], DeepSeek-V3 [25], Qwen2.5-72B-Instruct [26], and Mistral-Small-24B-Instruct [27]. All models used a temperature of 0.6, the maximum output length was configured to 4096 tokens, with default settings for all other parameters.

### D. Representative Multimorbidity Cases-based Evaluation

To assess the system's therapy recommendation capabilities, we conducted an evaluation using four clinical cases involving multimorbidity interactions. These cases are derived from a benchmark dataset for multimorbidity decision support, and were prepared by experienced clinicians [28], providing valuable insights into the system's potential applicability in real-world clinical settings. The 4 cases are summarized below. They included clinician-curated gold standard based on evidence-based guidelines.

The entire conflict resolution workflow was applied to these four cases. The final treatment plans generated by the system were then compared with the corresponding gold standard. Other generated advices were discussed with the clinician to determine true from wrong advices (see III- E).

*Case 1: TIA-DU-Osteoporosis*

**Clinical scenario:** A 76-year-old female with Transient Ischemic Attack (TIA) and Duodenal Ulcer (DU) required antithrombotic therapy (aspirin) for secondary prevention of stroke. Aspirin-induced DU necessitated proton-pump inhibitor (PPI) use, exacerbating osteoporosis—a therapeutic conflict. The gold standard solution prioritized replacing aspirin with clopidogrel (reducing DU risk) and discontinuing PPI, resolving both conflicts (Preferred Option). Alternatives included maintaining aspirin/PPI (Option 1: continued osteoporosis risk) or stopping PPI (Option 2: DU recurrence risk). Note that we used the benchmark case as is, but accepted correct solutions that leave the PPI, as the current clinical guidelines recommend its continuation.

*Case 2: CKD-HTN-AFib*

**Clinical scenario:** A 70-year-old male with non-proteinuric CKD and hypertension (HTN) (treated with ACE inhibitor, CCB, and diuretic) developed highly symptomatic persistent atrial fibrillation (AFib). Existing low-dose aspirin for CVD prevention conflicted with AFib management requiring long-term anticoagulation. Preferred option is to replace aspirin with a DOAC (e.g., apixaban) to reduce CVD risk and align with patient preference to avoid warfarin/INR monitoring. Initiate sodium channel blocker (SCB, propafenone) for rhythm control, preferring it over potassium channel blocker (PCB). BBs are avoided due to potential adverse interactions with non-dihydropyridine CCB (diltiazem).

*Case 3: VTE-UTI*

**Clinical scenario:** A 67-year-old female on long-term warfarin (5 mg/day) for Venous thromboembolism prevention due to a prosthetic aortic heart valve, developed recurrent urinary tract infection (UTI), confirmed by lab results. TMP/SMX (Bactrim) was prescribed, as the bacteria was defined as sensitive to this antiobiotic. The preferred option is either to switch the antibiotic or to mitigate the known interaction between TMP/SMX and warfarin, warfarin dose was slightly reduced by 10-20%. INR monitoring is initiated on day 3-5 of antibiotic therapy, and continued close monitoring assuring goal INR is achieved.

*Case 4: Stent and Surgery*

**Clinical scenario:** A 73-year-old male, two months post–MI and drug-eluting stent (DES) placement, was on dual antiplatelet therapy (DAPT: aspirin + clopidogrel). A newly diagnosed lung mass required urgent surgery, creating a critical conflict between maintaining antiplatelet therapy to prevent stent thrombosis and suspending it to reduce perioperative bleeding risk. Preferred option is to suspend clopidogrel 5 days pre-op. Initiate bridging therapy with tirofiban (or cangrelor) continuous intravenous drip 48 hours after clopidogrel cessation, stop tirofiban (or cangrelor) 4-6 hours before surgery. Clopidogrel is restarted within 24 hours post-surgery or later as soon as possible according to the post-surgical bleeding risk, with a 300 mg loading dose. Aspirin is maintained throughout.

*E. Evaluation metrics*

For each benchmark case study result, we extracted and organized relevant items, and compared them against the corresponding gold standard. The gold standard for each case included one or more management option sets. "A management option set, OS, is a collection of non-conflicting management plans (goals and the associated actions), together with their explanations, that a user can choose to manage a multimorbidy patient" [4]. We classify each item according to their different level of accuracy. For gold standard related Option Set's actions, we used the following classifications:

(1) Exact match or alternative correct (True positives, TP): the action is either exactly matched or replaced by a clinically acceptable alternative. (2) Omission (False negatives, FN): the action is not mentioned, or mentioned but with partial omissions. (3) Imprecise match (FN, but due to partial match, is scored as half the weight of the other scores) -- the action is mentioned but with inaccurate details. (4) Error (False positives, FP-wrong): the action is suggested only by the LLM, but is incorrect or contradicts medical knowledge.

For "Other Actions" that the LLM system mentioned but were not part of the gold standard, we consulted with a clinical expert (coauthor IH) to distinguish two cases: (1) The action is correct (FP-correct, yet is trivial and not part of the gold standard so no points were given or deducted for these FP). (2) The action is wrong for this case (FP-wrong).

After completing the classification, we counted across all option sets for a case study, the number of actions that were correctly matched (TP), actions that were in the gold standard but were missed by the LLM (FN), and additional actions that were not in the gold standard (FP-wrong), (for details, please refer to Github). From these we computed correctness (precision) and completeness (recall) to evaluate the gap between LLM and the gold standard: Correctness = TP/(TP+(FP-wrong)), Completeness = TP/(TP+FN). We also included a metric that assess whether the preferred option set was fully identified by the LLM.

In addition, we also computed DDI Ratio, Contraindication Ratio, and Medication Ratio, as defined below, to assess the size and direction of changes to DDI and medication number introduced by LLM-generated medication plans. The formulas below are similar to Hazard Ratio and produce numbers bounded by [-1, 1].

DDI Ratio (DDI-R) = #DDIs in revised plan / #DDIs in original plan. Note that DDI-R<1 is better

Contraindications Ratio (CR) = # Conflicts in revised plan / # Conflicts in original plan, where conflicts refer to those condition-specific contraindications (not DDIs) that may bring risks (Note that CR<1 is better)

Met Goals Ratio (GR) = Ratio of met goals in revised plan / Ratio of met goals in original plan (GR>1 is better)

Medication Ratio (MR) = # Medications in revised plan / # Medications in original plan. Note that MR<1 is better

In addition, four clinical experts were invited to independently assess the MAS process. Each expert rated the system using a five-point Likert scale (1 = lowest, 5 = highest), focusing on three components:

(1) Explainability, assessing whether the rationale behind identified and mitigated conflict was clear to clinicians.

TABLE I. STATISTICAL RESULTS FOR CASE 1-4

| Case | Metrics | Pure | | | | Single-agent | | | | Multi-agent | | | |
|---|---|---|---|---|---|---|---|---|---|---|---|---|---|
| | | C1: Qwen2.5 | C2: GPT-4o | C3: DeepSeek-V3 | C4: Mistral | C5: Qwen2.5 | C6: GPT-4o | C7: DeepSeek-V3 | C8: Mistral | C9: Qwen2.5 | C10: GPT-4o | C11: DeepSeek-V3 | C12: Mistral |
| Case 1 | Is the preferred option set included? | No | No | Same as C2 | No, C1&C2 | Yes | Same as C4 | Same as C1 | | Same as C5 | Same as C1 | | Same as C4 |
| | Correctness | 2/2 | 3/3 | | 5/5 | 2/2 | | | | | | | |
| | Completeness | 2/7 | 3/7 | | 5/7 | 2/7 | | | | | | | |
| | DDI Ratio | 0/0 | 0/0 | | 0/0 | 0/0 | | | | | | | |
| | Contraindication Ratio | 1/2 | 2/2 | | 1.5/2 | 0/2 | | | | | | | |
| | Medication Ratio | 3/2 | 3/2 | | 3/2 | 2/2 | | | | | | | |
| | Met-goal Ratio | 1.5 | 1.5 | | 1.5 | 1.5 | | | | | | | |
| Case 2 | Correctness | 6/7 | 5/7 | 5/8 | 7/9 | 5/5 | 4/5 | 5/6 | 4/7 | Same as C5 | 4/5 | Same as C5 | 5/6 |
| | Completeness | 5/6 | 4/6 | 4/6 | 5/6 | 5/6 | 4/6 | 5/6 | 4/6 | | 4/6 | | 5/6 |
| | DDI Ratio | 1/0 | 2/0 | 2/0 | 2/0 | 0/0 | 0/0 | 1/0 | 2/0 | | 1/0 | | 1/0 |
| | Contraindication Ratio | 1/1 | 1/1 | 1/1 | 1/1 | 0/1 | 0/1 | 0/1 | 1/1 | | 0/1 | | 0/1 |
| | Medication Ratio | 7/5 | 7/5 | 7/5 | 8/5 | 5/5 | 5/5 | 6/5 | 6/5 | | 5/5 | | 6/5 |
| | Met-goal Ratio | 0.83 | 0.5 | 0.83 | 1 | 1 | 0.83 | 1 | 0.5 | | 0.83 | | 1 |
| Case 3 | Correctness | 2.5/3 | Same as C1 | | | 2/2 | Same as C5 | | | | | | |
| | Completeness | 2.5/5 | | | | 2/2 | | | | | | | |
| | DDI Ratio | 1/1 | | | | 0/1 | | | | | | | |
| | Contraindication Ratio | 0/0 | | | | 0/0 | | | | | | | |
| | Medication Ratio | 2/2 | | | | 2/2 | | | | | | | |
| | Met-goal Ratio | 1 | | | | 1.5 | | | | | | | |
| Case 4 | Correctness | 3/3 | 4/4 | 4/4 | Same as C1 | 3/4 | 3/3 | 4/5 | Same as C1 | 3.5/5 | 4/4 | 4/5 | 6.5/7 |
| | Completeness | 3/6 | 4/6 | 4/6 | | 3/6 | 2/6 | 4/6 | | 3.5/6 | 4/6 | 4/6 | 5.5/6 |
| | DDI Ratio | 0/0 | 0/0 | 0/0 | | 0/0 | 0/0 | 0/0 | | 0/0 | 0/0 | 0/0 | 0/0 |
| | Contraindication Ratio | 0/1 | 0/1 | 0/1 | | 0/1 | 0/1 | 0/1 | | 0/1 | 0/1 | 0/1 | 0/1 |
| | Medication Ratio | 1.5/2 | 2/2 | 2/2 | | 2/2 | 2/2 | 2/2 | | 2/2 | 2/2 | 2/2 | 2/2 |
| | Met-goal Ratio | 1.5 | 2 | 2 | | 1.5 | 2 | 2 | | 1.5 | 2 | 1.5 | 2 |

(2) For conflict allocation in the multi-agent versions, experts rated (a) Reasonableness, reflecting whether the assignment of each conflict to the specialist agents was appropriate; and (b) Efficiency, ie, how well the system avoided redundant consultations and ensured that each conflict was handled by the most appropriate agent with minimal delay.

For each case, all expert scores were inspected. Where there was high lack of agreement among the experts, they discussed until consensus was reached. When there was reasonable agreement, the scores were averaged. The aggregated results were visualized using radar charts.

## IV. RESULTS AND ANALYSIS

The Github appendix provides the complete results. The calculated metrics are shown above (Table I). These with green color indicates performance comparable to the best gold standard, while red represents inferior performance. Note that only Case 1 had more than one option set.

### A. Case 1: TIA-DU-Osteoporosis

In this case study, there were three option sets, one of which was the preferred one; it included a replacement of medication (aspirin) and meeting of all clinical goals, which were reduced from 3 to 2. Replacement decisions, whether in the MAS or in a single LLM, are less common; only the Qwen2.5 based Single-agent and Multi-agent systems were able to propose the preferred option and scored well on all but completeness There was no difference between the Single-agent and Multi-agent systems, except that the multi-agent system was able to offer more additional (Other) suggestions.

As shown in Table I, all groups achieved perfect scores in correctness (they did not suggest FP-wrong). However, their completeness was not so good, as each successfully identified only some of the medications (for the multiple clinical goals) for only some of the option sets. Medication number increased mainly due to the osteoporosis management.

### B. Case 2: CKD-HTN-AFib

This Case was not solved well by any LLM version, and only 3 of 4 clinical goals were met (in initial and final recommendations). We refer the readers to the Github for more details.

### C. Case 3: VTE-UTI

As presented, the Pure group yields the lowest performance with both correctness and completeness lower than 1. In contrast, the other two groups proposed clinically appropriate alternatives that resolved the DDI (DDI-R = 0), along with meet all clinical goals. In this Case, both the Single-agent and Multi-agent groups recommended to use nitrofurantoin to replace TMP/SMX, which is clinically acceptable alternative. And the Pure group failed to specify a dosage adjustment. In terms of INR monitoring frequency, the alternative plan proposed by the Single-agent and Multi-agent, though not aligned with the gold standard, is reasonable given that it does not require frequent INR monitoring.

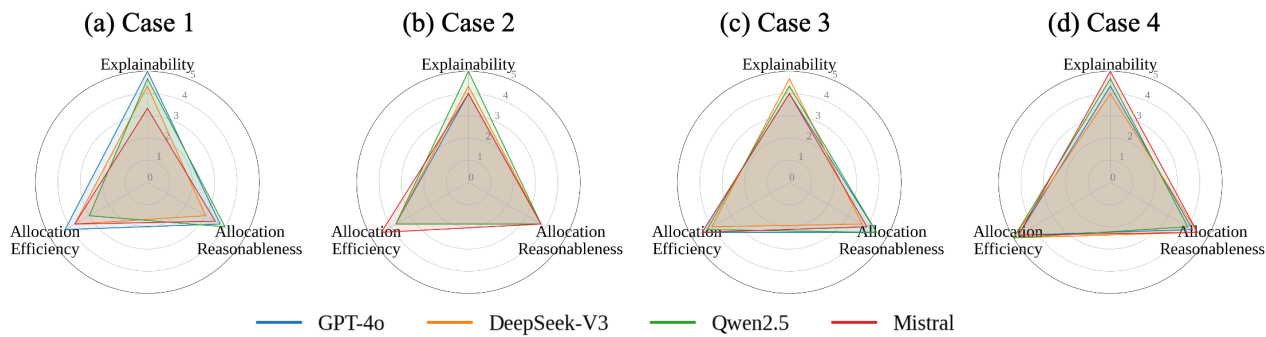

Fig. 2. Expert evaluation scores for MAS across four cases and 4 models.

In this case, one main DDI is included (TMP/SMX + warfarin), and only can be addressed when replacing TMP/SMX with an alternative antibiotic.

*D. Case 4: Stent and Surgery*

As shown in Table I, all Pure LLMs and the GPT-4o MAS performed well in achieving clinical goals and correctness, with completeness of 4/6.

Clopidogrel suspension 5 days prior to surgery was correctly recommended by 4 out of 4 Pure models, 2 out of 4 single-agent models, and 3 out of 4 multi-agent models. Option Set actions regarding bridging therapy were primarily derived from the literature [30] rather than clinical guidelines (this benchmark case study material included a clinical guideline, its update, and supporting literature [30-32]). Notably, the Pure model failed to provide any relevant information on this issue, while the MAS offered more comprehensive insights. Interestingly, only the Mistral-based MAS performed well in this issue. More details in Github.

*E. Physician Evaluation Results*

Clinical expert evaluation targeted the appropriateness of the multidisciplinary team and quality of the explanations. Results, visualized in Fig. 2, revealed that model performance varied across clinical cases and evaluation dimensions. No single model consistently outperformed others across all cases. In terms of reasonableness of the MDT the best score (for any model and any case study ranged in [3.8, 4.5] and the worst in [3.0, 4.0]. For efficiency, the best ranged in [4.3, 5.0] and the worst in [3.0, 4.8]. This means that the clinical experts found the MDT composition and size to be very high for some models and case studies but only fair for others. Qwen2.5 and Mistral performed well in the dimension of allocation reasonableness and efficiency, respectively, each receiving higher or comparable scores in 3 out of 4 cases compared to other models.

For explainability, the best scores were in the range of [4.7, 5.0] and the worst score range was [3.3, 4.0]. Mistral demonstrated the best score for explainability in case 4 (5.00 ± 0.00), but the lowest score in case 1 (3.33 ± 0.58). This is primarily due to differences in explanation structure. In Case 1, the other three models followed a structured format of goal, intervention, and rationale, clearly explaining the trade-offs between clinical goals and potential conflicts. In contrast, Mistral only presented the goal and intervention without further explanation, resulting in a lower explainability score. Conversely, Mistral received the top score in Case 4 by adopting a clinically aligned reasoning process, organizing its explanation into stages (preoperative, intraoperative, postoperative). Although GPT-4o also followed this structure, it lacked details in certain clinical parameters (e.g., drug dosages), reducing the clinical operability of its plan compared to the Mistral's comprehensive explanation.

## V. DISCUSSION

This study introduces LLM systems based on a single GP agent or MAS MDT, that simulate decision-making for medication-related patient management recommendations, and investigates its ability to detect and mitigate conflicts. A **key contribution** is the development of a multi-step workflow that mirrors the decision-making logic as done in real life by clinicians, enabling LLMs to detect medication conflicts and propose improved plans.

Two key questions could be answered by our analysis. First, do we need an MDT, or is a single GP agent enough, and is intermediate reasoning of the LLMs beneficial? As seen by the blue rectangles in Table 1, that outline the best solutions that address all clinical goals, it seems that with current LLMs, MDTs are not necessarily needed since the single agent model with intermediate reasoning can perform as well as the multi agent.

The second question is whether LLMs could be beneficial for decision support. In this experiment, we did not evaluate the systems support to clinical decision making of clinician users. We note that most of the errors of the LLMs were errors of omission and not of commission. Interestingly, these are also the type of errors made by clinicians when they are not using decision support systems to manage multimorbidity patient cases [4]. Therefore, the value of LLM-based systems, in the current stage of LLMs does not seem to be high, but this needs to be evaluated again in the future as LLMs evolve.

Evaluation on the benchmark cases showed that compared with Single-agent, the MAS has promise in improving certain cases by reducing conflicts while maintaining medication number in an appropriate level, and it is more closely matched with real-world expert judgment for multimorbidity patients.

Results also revealed that model performance varied across clinical cases and evaluation dimensions. No single model consistently outperformed others across all cases.

**A second contribution** of our study is the creation of a more interpretable evaluation strategy based on realistic, expert-annotated benchmark cases [28], rather than aggregate metrics (i.e., completeness and correctness) alone. Evaluation should extend beyond conventional accuracy metrics.

Realistic cases with structured outputs and clinical benchmarks are both essential for meaningful validation [29]. We introduced a more specific improvement ratio that compares the LLM-proposed treatment plan with the original plan in terms of number of conflicts, medication burden, and extent of clinical goals met. These ratios are analogous to a hazard ratio. We also proposed metrics for clinicians evaluating whether the system forms appropriate MDT for conflict resolution. Specifically, experts rated the reasonableness and efficiency of conflict allocation on a five-point Likert scale, reflecting how well the system assigned conflicts. Our findings underscore the value of using expert-annotated benchmark cases that reflect the complexity of real-world decision-making.

However, several limitations were identified. First, while some LLM agents (Qwen2.5, case 1) were capable of recommending appropriate substitutions, we observed that others tended to recommend adding or removing medications but struggled to suggest suitable alternatives, which is critical for minimizing polypharmacy and conflicts while preserving coverage of clinical goals with indicated treatment options.

Second, such MAS' performance heavily relies on manually crafted prompts, that explicitly suggest drug replacements or considering the use of clinical guidelines and additional studies (without pointing to specific references).

A third limitation concerns the lack of references to important clinical studies or guidelines from the literature. In solving Case 4, the Mistral model suggested tirofiban and considered the timing related to suspension of clopidogrel. While this might initially suggest better exposure to relevant literature, the model itself denied to have the access to the specific paper when queried, suggesting that its good performance may due to the knowledge acquired during pretrain rather than explicit citation or retrieval; it seems that the multi-agent approach was able to extract more accurate information through agent-to-agent dialogue. Alternatively, it may have implicitly learned from the content during pretrain, even if it did not acknowledge it. Since LLMs generate outputs based on probabilistic token predictions rather than explicit database retrieval, a model's statement of unawareness cannot be equated with true ignorance. Conversely, other models such as GPT-4o, which claims to have been last updated in 2023, may have had access to the article during pretrain. However, due to limitations in the prompt, such as the absence of key terms, the model failed to retrieve or utilize the relevant information, resulting in only moderate performance. These observations underscore a key limitation, even when a model has been exposed to relevant sources, its ability to leverage them depends critically on how the prompt is framed. On the other hand, a model that does not explicitly reference a source may still approximate accurate outputs by generalizing from similar information in its training data. To enhance such performance, reinforcement learning could be employed to optimize prompts, which, in turn, may compromise their clinical relevance, highlighting the trade-off between optimization and interpretability [36].

Similarly, we also found that LLMs rarely cite clinical guidelines or explain dosing decisions without structured prompting, underscoring the importance of incorporating retrieval-augmented generation (RAG) frameworks or leveraging medical-specialized foundation models, which can inject high-quality, evidence-grounded knowledge into the reasoning process and improve clinical reliability [33].

Fourth, the bench mark was not designed with MDT focus. Although these cases involve multimorbidity patients, some may not fully reflect the complexity (e.g., case 1 and case 3), limiting the extent to which the advantages of multi-agent collaboration can bring.

Another critical limitation observed is that the LLM-generated recommendations didn't include all valid therapeutic options, regardless of single-agent or multi-agent setup (case 1). A good CDSS should not only identify one preferred solution, but rather present the full range of appropriate options, distinguishing between preferred and less-preferred choices and providing rationales, so as to support the clinician in selecting the most appropriate choice for the patient. This indicates moving from single-best-answer generation toward multi-option is a promising direction for future LLM-based CDSS development.

Based on our experiments, we summarize additional key lessons for future related works:

(1) Small, conflict-targeted MDTs improve clarity and realism in decision support workflows. Initially (results not presented), we adopted a strategy in which an MDT was convened at the start of the process—immediately after the GP reviewed the patient's condition, without focusing on conflicts. Each specialist proposed an independent medication plan, which was then merged into a composite plan for conflict detection. However, this approach often led to the formation of overly large MDTs, diverging from typical real-world practice, and introduced unnecessary or redundant medications, increasing the patient's medication burden and exacerbating the risk of conflicts. Hence the final MDT workflow assembles the MDT based on detected conflicts.

(2) MAS, via agent-to-agent dialogue, captured a broader scope of relevant clinical information, which in many cases yielded a more comprehensive clinical picture. However, this strength comes with trade-offs. The collaborative process occasionally introduced inaccurate details, increasing the risk of clinical errors.

(3) Structured output formats, such as explicit annotations of clinical goals, actions, and rationales, can enable more granular and interpretable evaluations. Future benchmarks should incorporate both outcome-oriented and process-oriented criteria to support a more holistic assessment of system performance.

(4) MAS performance is context-dependent and may benefit from adaptive model selection. No single base model demonstrated consistent superiority across all cases and evaluation dimensions. Performance varied depending on the nature of the clinical scenario and the type of task. Future systems may improve reliability by dynamically selecting or prioritizing models based on their demonstrated strengths for each clinical task [34, 35].

ACKNOWLEDGMENT

We also thank clinicians Mengjia Chen and Xiangjie Sun from The First Affiliated Hospital, Zhejiang University School of Medicine, for their valuable scoring support.

REFERENCES

[1] W. V. Woensel, S. W. Tu, Michalowski W, et al., "A community-of-practice-based evaluation methodology for knowledge intensive computational methods and its application to multimorbidity decision support," *J Biomed Inform.*, 142:104395, 2023.

[2] L. Piovesan, P. Terenziani, and G. Molino, "GLARE-SSCPM: an intelligent system to support the treatment of comorbid patients," *IEEE Intelligent Systems*, 33(6):37-46, 2018.

[3] W. V. Woensel, S. S. R. Abidi, and S. R. Abidi, "Decision support for comorbid conditions via execution-time integration of clinical guidelines using transaction-based semantics and temporal planning," *Artificial Intelligence in Medicine*, 118:102127, 2021.

[4] A. Kogan, M. Peleg, and S. W. Tu, "Towards a goal-oriented methodology for clinical-guideline-based management recommendations for patients with multimorbidity: GoCom and its preliminary evaluation," *J Biomed Inform.*, 112:103587, 2020.

[5] O. O. Ogbuagu, A. O. Mbata, O. D. Balogun, O. Oladapo, O. O. Ojo, and M. Muonde, "Artificial intelligence in clinical pharmacy: enhancing drug safety, adherence, and patient-centered care," *Internaltional Journal of Multidisciplinary Research and Grouth Evaluation*, 4(1):814-822, 2023.

[6] G. Lebedev, E. Fartushnyi, I. Fartushnyi, et al., "Technology of Supporting Medical Decision-Making Using Evidence-Based Medicine and Artificial Intelligence," *Procedia Computer Science*, 176:1703-1712, 2020.

[7] Y. Zhang, R. Chen, J. Tang, W. F. Stewart, and J. Sun, "Leap: Learning to prescribe effective and safe treatment combinations for multimorbidity," *Proceedings of the 23rd ACM SIGKDD International Conference on Knowledge Discovery and Data Mining*, 1315-1324, 2017.

[8] C. Yang, C. Xiao, F. Ma, L. Glass, and J. Sun, "SafeDrug: Dual Molecular Graph Encoders for Recommending Effective and Safe Drug Combinations," *Proceedings of the Thirtieth International Joint Conference on Artificial Intelligence (IJCAI-21)*, 3735-3741, 2021.

[9] F. Gong, M. Wang, H. Wang, S. Wang, and M. Liu, "SMR: Medical Knowledge Graph Embedding for Safe Medicine Recommendation," *Big Data Research*, 23:100174, 2021.

[10] Elsevier, "Attitudes toward AI," Jul. 9, 2024. [Online]. Available: https://www.elsevier.com/insights/attitudes-toward-ai.

[11] M. C. S. Menezes, A. F. Hoffmann, A. L. M. Tan, et al., "The potential of Generative Pre-trained Transformer 4 (GPT-4) to analyse medical notes in three different languages: a retrospective model-evaluation study," *Lancet Digit Health*, 7:e35-e43, 2025.

[12] M. Tordjman, Z. Liu, M. Yuce, et al., "Comparative benchmarking of the DeepSeek large language model on medical tasks and clinical reasoning," *Nat Med*, 2025.

[13] J. C. L. Ong, M. H. Chen, N. Ng, et al., "A scoping review on generative AI and large language models in mitigating medication related harm," *npj Digit Med*, 8:182, 2025.

[14] P. Festor, Y. Jia, A. C. Gordon, A. A. Faisal, I. Habli, and M. Komorowski, "Assuring the safety of AI-based clinical decision support systems: a case study of the AI Clinician for sepsis treatment," *BMJ Health Care Inform*, 29(1):e100549, 2022.

[15] Y. Si, Y. Yang, X. Wang, et al., "Quality and Accountability of ChatGPT in Health Care in Low- and Middle-Income Countries: Simulated Patient Study," *J Med Internet Res*, 26:e56121, 2024.

[16] Z. Fan, L. Wei, J. Tang, et al., "AI Hospital: Benchmarking Large Language Models in a Multi-agent Medical Interaction Simulator," In *Proceedings of the 31st International Conference on Computational Linguistics*, 10183-10213, 2025.

[17] Z. Wang, Y. Zhu, H. Zhao, et al., "ColaCare: Enhancing Electronic Health Record Modeling through Large Language Model-Driven Multi-Agent Collaboration," In *Proceedings of the ACM on Web Conference*, 2250-2261, 2025.

[18] J. Wei, D. Yang, Y. Li, et al., "Medaide: Towards an omni medical aide via specialized llm-based multi-agent collaboration," 2024, *arXiv: 2410.12532*. [Online]. Available: https://arxiv.org/abs/2410.12532

[19] Y. Kim, C. Park, H. Jeong, et al., "MDAgents: An Adaptive Collaboration of LLMs for Medical Decision-Making," In *The Thirty-eighth Annual Conference on Neural Information Processing Systems*, 2024.

[20] X. Chen, H. Yi, M. You, et al., "Enhancing diagnostic capability with multi-agents conversational large language models," *npj Digit Med*, 8:159, 2025.

[21] C. J. Charlesworth, E. Smit, D. S. Lee, F. Alramadhan, and M. C. Odden, "Polypharmacy among adults aged 65 years and older in the United States: 1988–2010," *J Gerontol A Biol Sci Med Sci*, 70:989-995, 2015.

[22] M. Zitnik, M. Agrawal, and J. Leskovec, "Modeling polypharmacy side effects with graph convolutional networks," *Bioinformatics*, 34:i457-i466, 2018.

[23] X. Tang, A. Zou, Z. Zhang, et al., "MedAgents: Large Language Models as Collaborators for Zero-shot Medical Reasoning," In *Findings of the Association for Computational Linguistics: ACL*, 599-621, 2024.

[24] OpenAI, et al., "GPT-4o System Card," 2024, *arXiv: 2410.21276*. [Online]. Available: https://arxiv.org/abs/2410.21276

[25] DeepSeek-AI, et al., "DeepSeek-V3 Technical Report," 2024, *arXiv: 2412.19437*. [Online]. Availavle: https://arxiv.org/abs/2412.19437

[26] Qwen, et al., "Qwen2.5 Technical Report," 2024, *arXiv: 2412.15115*. [Online]. Availavle: https://arxiv.org/abs/2412.15115

[27] Mistral AI team, "Mistral small 3," Jan. 30, 2025. [Online]. Available: https://mistral.ai/news/mistral- small-3/

[28] D. O'Sullivan, W. V. Woensel, S. Wilk, et al., "Towards a framework for comparing functionalities of multimorbidity clinical decision support: A literature-based feature set and benchmark cases," *AMIA Annu Symp Proc*, 2021:920-929, 2022.

[29] Stanford university, "Holistic Evaluation of Large Language Models for Medical Applications," Feb. 28, 2025. [Online]. Available: https://hai.stanford.edu/news/holistic-evaluation-of-large-language-models-for-medical-applications

[30] S. Savonitto, M. D'Urbano, M. Caracciolo, et al., "Urgent surgery in patients with a recently implanted coronary drug-eluting stent: a phase II study of 'bridging'antiplatelet therapy with tirofiban during temporary withdrawal of clopidogrel," *British journal of anaesthesia*, 104(3):285-291, 2010.

[31] M. Valgimigli, H. Bueno, R. A. Byrne, et al., "2017 ESC focused update on dual antiplatelet therapy in coronary artery disease developed in collaboration with EACTS," *European Heart Journal*, 39:213-254, 2018.

[32] P. G. Chassot, C. Marcucci, A. Delabays, and D. R. Spahn, "Perioperative antiplatelet therapy," *Am Fam Physician.*, 82(12):1484-1489, 2010.

[33] Y. Kim, H. Jeong, C. Park C, et al., "Tiered Agentic Oversight: A Hierarchical Multi-Agent System for AI Safety in Healthcare," 2025, *arXiv: 2506.12482*. [Online]. Available: https://arxiv.org/abs/2506.12482

[34] H. Zhang, Z. Cui, X. Wang, et al., "If Multi-Agent Debate is the Answer, What is the Question?" 2025, *arXiv: 2502.08788*. [Online]. Available: https://export.arxiv.org/abs/2502.08788

[35] P. Xia, J. Wang, Y. Peng, et al., "MMedAgent-RL: Optimizing Multi-Agent Collaboration for Multimodal Medical Reasoning," 2025, *arXiv: 2506.00555*. [Online]. Available: https://arxiv.org/abs/2506.00555

[36] M. Deng, J. Wang, C. P. Hsieh, et al., "RLPrompt: Optimizing Discrete Text Prompts with Reinforcement Learning," In *Proceedings of the 2022 Conference on Empirical Methods in Natural Language Processing*, 3369-3391, 2022.